\documentclass{article}
\usepackage{spconf,amsmath,graphicx}
\usepackage{amsfonts}
\usepackage{multirow}
\usepackage{subfigure}
\usepackage[colorlinks,linkcolor=red]{hyperref}

\title{DEFORMABLE SPATIAL PROPAGATION NETWORKS FOR DEPTH COMPLETION}
\name{Zheyuan Xu$^1$, Hongche Yin$^1$, Jian Yao\sthanks{Corresponding Author}$^{1,2}$}
\address{$^1$School of Remote Sensing and Information Engineering, Wuhan University, Wuhan, P.R. China\\
$^2$Shenzhen Jimuyida Technology Co., Ltd., Shenzhen, P.R. China\\
{$^*$Email: \href{mailto:jian.yao@whu.edu.cn}{jian.yao@whu.edu.cn} Web: \href{http://cvrs.whu.edu.cn/}{http://cvrs.whu.edu.cn/}}}
\begin{document}
%\ninept
%
\maketitle
\begin{abstract}
Depth completion has attracted extensive attention recently due to the development of autonomous driving, which aims to recover dense depth map from sparse depth measurements. Convolutional spatial propagation network (CSPN) is one of the state-of-the-art methods in this task, which adopt a linear propagation model to refine coarse depth maps with local context. However, the propagation of each pixel occurs in a fixed receptive field. This may not be the optimal for refinement since different pixel needs different local context. To tackle this issue, in this paper, we propose a deformable spatial propagation network (DSPN) to adaptively generates different receptive field and affinity matrix for each pixel. It allows the network obtain information with much fewer but more relevant pixels for propagation. Experimental results on KITTI depth completion benchmark demonstrate that our proposed method achieves the state-of-the-art performance.
\end{abstract}
\begin{keywords}
depth completion, deformable, spatial propagation, KITTI dataset
\end{keywords}
\section{Introduction}
\label{sec:intro}
Depth perception and estimation is fundamental in many applications, such as robotics, autonomous driving, augmented reality (AR) and 3D mapping. However, existing depth sensors produce depth maps with incomplete data. For example, LiDARs have limited scanlines and scan frequencies, and thus only provide sparse depth measurements. Therefore, the depth completion task, which estimates dense depth maps from sparse depth measurements has attracted extensive attention recently.

\begin{figure}[t]
  \centering
  \includegraphics[width=0.5\textwidth]{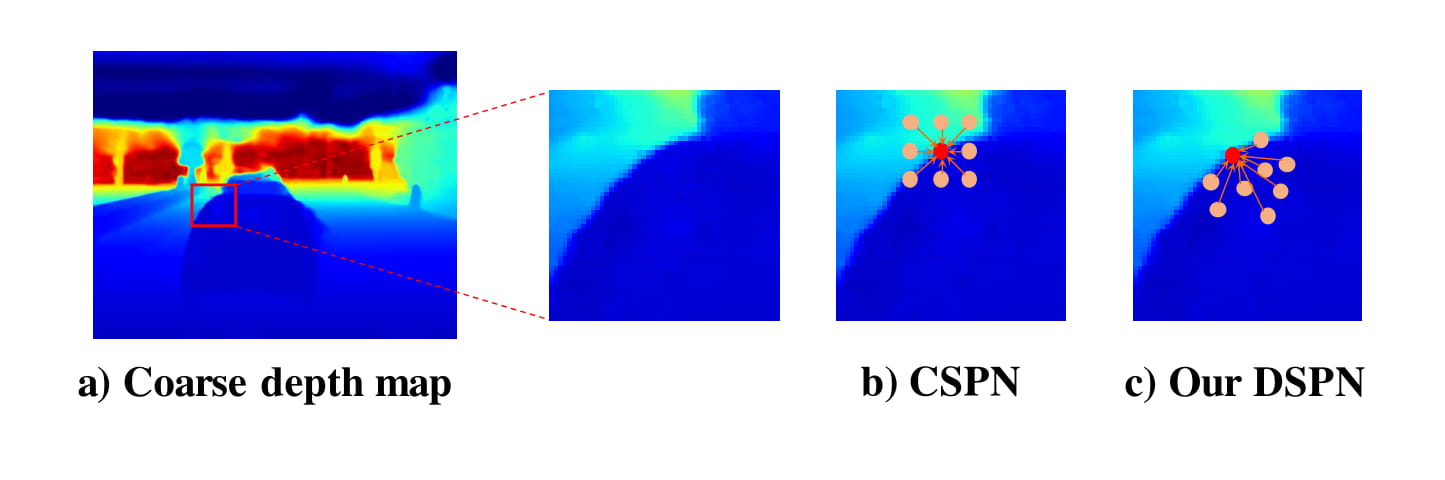}
  \label{motivation}
  \vspace{-0.8cm}
  \caption{a) The coarse depth map output from the prediction network. b) The propagation of each pixel in CSPN. c) The propagation of each pixel in our proposed DSPN. Compared with CSPN, in DSPN, each pixel obtains adaptive local context information in propagation.}
\end{figure}

The traditional methods are often based on interpolation and diffusion, which use corresponding RGB image as a guide to up-sample sparse point into dense depth image \cite{ferstl2013image}. Recently, since convolutional neural networks (CNNs) have achieved tremendous success in depth estimation tasks using monocular image \cite{li2018megadepth,fu2018deep,8804273,8803006}, deep learning based approaches have also become the mainstream in depth completion tasks, which take sparse depth maps (with/without RGB images) as input and adopt an encoder-decoder network to predict dense depth maps \cite{mal2018sparse,jaritz2018sparse,uhrig2017sparsity,eldesokey2019confidence,eldesokey2018propagating,qiu2019deeplidar,xu2019depth} and have achieved significant improvements. Nevertheless, the depth map directly predicted from these network is still blurry. To refine the coarse predicted depth, Cheng \textit{et al}.\ \cite{cheng2018depth,cheng2019learning} propose an efficient local linear propagation model named convolutional spatial propagation networks (CSPN), where the depth values at all pixels are updated simultaneously with a local convolutional context. However, its propagation of each pixel is performed in the same size receptive field. Intuitively, as shown in Fig.\ \ref{motivation}\ (b), this refinement of pixels at object boundaries may introduce irrelevant information. Thus, depth refinement of each pixel needs different local context. 

To address these drawbacks, in this paper, we attempt to provide each pixel with a different receptive field, e.g., Fig.\ \ref{motivation}\ (c). Inspired by \cite{dai2017deformable}, we propose a more flexible and effective approach, known as Deformable Spatial Propagation Network (DSPN). It learns adaptive offsets and affinity matrices in a data-driven manner for propagation. Specifically, offsets determines the receptive field of each pixel in propagation, and affinity matrices determine the effectiveness of pixels in such receptive field. Compared with CSPN, our proposed DSPN allows to use far fewer but more effective pixels to deal with depth refinement. 
Moreover, a confidence branch is introduced in our framework, which predicts confidence masks of sparse depth measurements to mitigate the effects of sensor noise.

\begin{figure*}[t]
  \centering
  \includegraphics[width=\textwidth]{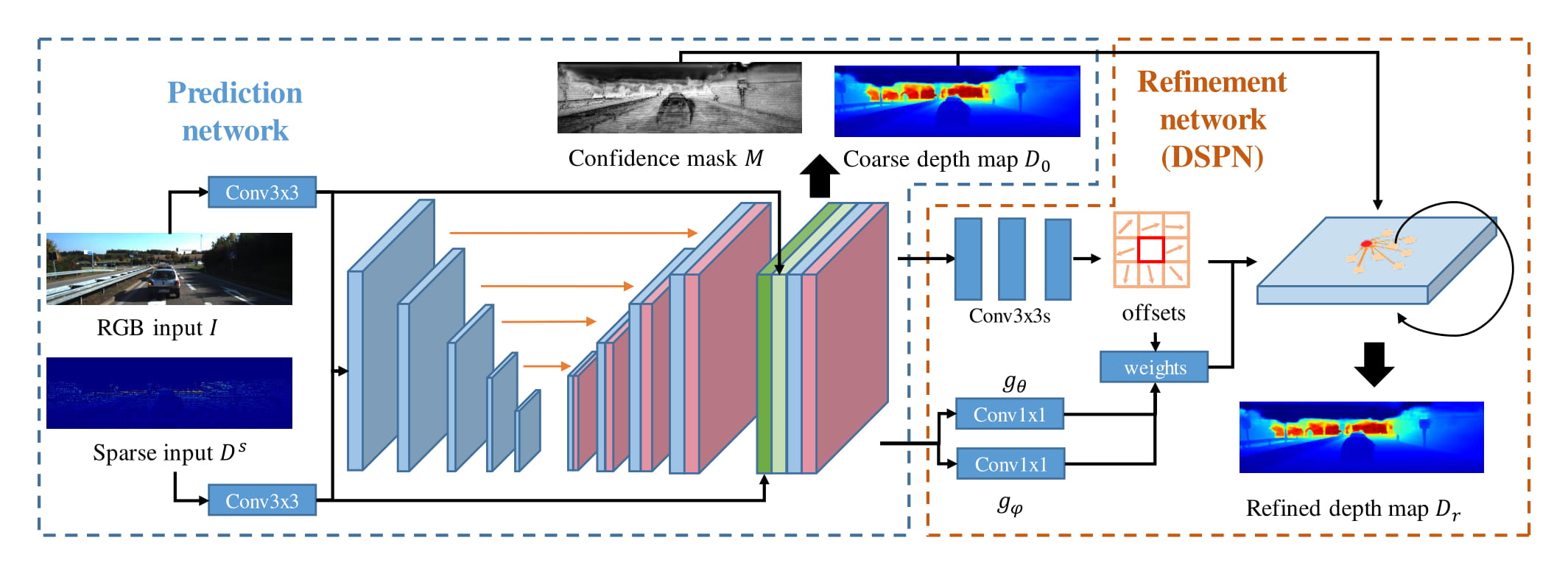}
  \caption{{\bf Overview of our proposed framework.} The prediction network takes sparse depth $D^s$ and RGB image $I$ as input, and predicts coarse depth $D_0$ and confidence map $M$. The refinement network (DSPN) first predicts offsets, and then updates the coarse depth map $D_0$ based on the above offsets, using Eq.\ \ref{eq:dspn1}. During the refinement, the propagation conduction depends on the similarity in features generated in prediction network (Eq.\ \ref{similarity}). Finally, the refined depth map $D_r$ is computed by Eq.\ \ref{update2} when the propagation is finished.}
  \label{pipline}
\end{figure*}

The main contributions in this paper mainly lie in three aspects: (1) We construct an end-to-end framework to produce dense depth maps from a sparse LiDAR depth and a monocular RGB image.(2) We propose deformable spatial propagation networks (DSPN) to refine depth maps, which is more efficient and flexible than CSPN in depth completion task. (3) The experimental results on the challenging KITTI depth completion benchmark \cite{uhrig2017sparsity} show that our model outperforms the state-of-the-art methods.

\section{Methodology}
In this section, we first briefly review the CSPN, and then illustrate the proposed deformable spatial propagation networks (DSPN) and the framework for depth completion.

\subsection{Review the CSPN}
Given the coarse predicted depth map $D_0$, CSPN \cite{cheng2018depth,cheng2019learning} iteratively generates a new depth map $D_t$. Without loss of generality, to follow their formulation, we embed depth to a hidden representation $H\in\mathbb{R}^{m\times n\times c}$, and the one step propagation of kernel size $k$ can be formulated as:
\begin{equation}
  \begin{aligned}
    H_{t+1}(x_i)=\kappa_{x_i}(x_i)\odot H_t(x_i)+\sum_{x_j\in\mathcal{N}_k(x_i)}\kappa_{x_i}(x_j)\odot H_t(x_j)\\
    \kappa_{x_i}(x_j)=\frac{\hat{\kappa}_{x_i}(x_j)}{\sum_{x_j\in\mathcal{N}}|\hat{\kappa}_{x_i}(x_j)|},\ \kappa_{x_i}(x_i)=1-\sum_{x_j\in\mathcal{N}}\kappa_{x_i}(x_j)
  \end{aligned}
  \label{eq:cspn}
\end{equation}
where $\mathcal{N}_k(x_i)$ is the set of neighborhood pixels, $\hat{\kappa}_x\in\mathbb{R}^{k\times k\times c}$ is the output from an affinity network, and $\odot$ denotes the element-wise product. During this propagation process, pixel $x_i$ receives information from surrounding pixels in neighborhood $\mathcal{N}_k(x_i)$. 

In depth completion, CSPN preserves the depth values provided by the depth sensor. Specifically, sparse depth map $D^s$ is also embedded to a hidden representation $H^s$ and a replacement operation is added after each step:
\begin{equation}
  H_{t+1}(x)=(1-m(x))H_{t+1}(x)+m(x) H^s(x)
  \label{update1}
\end{equation}
where $m(x)=I(D^s(x)>0)$ is an indicator for the availability of sparse depth $D^s$.

\subsection{Deformable Spatial Propagation Network}
\label{dspn}
In the convolutional spatial propagation network, the propagation on the depth maps occurs in a fixed local context, which may not be the best choice as explained in Sec.\ \ref{sec:intro}. To solve this problem, we propose a deformable spatial propagation network (DSPN). In detail, according to Eq.\ \ref{eq:cspn}, the update formulation of arbitrary pixel is actually decided by the affinity matrix and the pixels in its receptive field, i.e., $\hat{\kappa}_x$ and $\mathcal{N}_k(x)$ in CSPN. To make the receptive field adaptive, we use a offset estimator to produce a offset $\Delta p_n$ for each pixel $x_n\in\mathcal{N}_k(x)$, where the offset estimator consists of 3 convolutional layers. Thus each pixel will obtain an adaptive receptive field $\widetilde{\mathcal{N}}_k(x)=\{x_n+\Delta p_n|x_n\in\mathcal{N}_k(x)\}$. Then, the one step propagation in DSPN could be written as
\begin{equation}
  \begin{aligned}
    H_{t+1}(x_i)&=(1-\sum_{\widetilde{x}_j\in\widetilde{\mathcal{N}}_k(x_i)}\omega(x_i,\widetilde{x}_j))\odot H_t(x_i)\\
    &+\sum_{\widetilde{x}_j\in\widetilde{\mathcal{N}}_k(x_i)}\omega(x_i,\widetilde{x}_j)\odot H_t(\widetilde{x}_j)
  \end{aligned}
  \label{eq:dspn1}
\end{equation}
where $H_t(\widetilde{x}_j)$ is the bilinear interpolation of depth map $H_t$ at position $\widetilde{x}_j$, and $\omega(x_i,\widetilde{x}_j)$ measures the affinity between $x_i$ and $\widetilde{x}_j$. Different from CSPN directly generating affinity matrices through a network, we learn the affinity matrices adaptively by measuring the similarity between $x_i$ and $\widetilde{x}_j$ in the high-dimension feature space. 

Specifically, we take the feature map $F$ output from the last layer of the prediction network to model the similarity between $x_i$ and $\widetilde{x}_j\in\widetilde{\mathcal{N}}_k(x_i)$. 
Similar to \cite{vaswani2017attention}, two learnable matrices are used to embed features, then dot product is adopted to measure the similarity of two pixels $x_i$ and $\widetilde{x}_j$ in an embedding space. Formally, the similarity is computed as follow:
\begin{equation}
  \omega(x_i, \widetilde{x}_j)=\frac{1}{Z(x_i)}exp(\frac{g_\theta(F(x_i))^Tg_\phi(F(\widetilde{x}_j))}{\sqrt{d_F}})
  \label{similarity}
\end{equation}
where $F(\widetilde{x}_j)$ is the bilinear interpolation of feature map $F$ at position $\widetilde{x}_j$. $g_\theta$ and $g_\phi$ are two different learnable matrices, which makes the propagation asymmetric, and such asymmetric provides more flexibility for propagation. $d_F$ is the dimension of features, denotes as the scaling factor. $Z(x_i)=\sum_{x\in\widetilde{\mathcal{N}_k}(x_i)\cup\{x_i\}}exp(g_\theta(F(x_i))^Tg_\phi(F(x))/\sqrt{d_F})$ is a normalization term.

Moreover, in depth completion task, to mitigate the measurement errors caused by noises in practical LiDAR, we introduce a confidence branch to predict a continuous confidence mask $M$ for sparse depth. During training, we follow \cite{xu2019depth} to use a function to model the ground-truth of confidence map:
\begin{equation}
  M^*(x)=m(x)\cdot exp(-\frac{|D^*(x)-D^s(x)|}{\gamma})
\end{equation} 
where $\gamma$ is a tolerance factor and $D^*$ is the ground-truth of depth map. Thus, the update function Eq.\ \ref{update1} will be transformed into:
\begin{equation}
  \begin{aligned}
    H_{t+1}(x)&=(1-m(x)M(x))H_{t+1}(x)\\
    &+m(x)M(x)H^s(x)
  \end{aligned}
  \label{update2}
\end{equation}

\subsection{Network Architecture}\label{net}
As illustrated in Fig.\ \ref{pipline}, the whole network contains a prediction network and a refinement network. The prediction network takes sparse depth $D^s$ and corresponding RGB image $I$ as input, and output coarse depth map $D_0$ and confidence map $M$. In detail, we follow an encoder-decoder paradigm to use a ResNet-34 variant as the encoder and cascaded upsample layers as decoder. In encoder, sparse depth and RGB image are convolved separately first, then concatenated and fed into ResNet-34 to extract high-level features. In decoder, each upsample layer contains a bilinear interpolation operation and a convolutional operation. In particular, we add skip connections similar to U-Net \cite{ronneberger2015u}, i.e., concatenating the features from encoder to decoder. The refinement network is our proposed DSPN introduced in Sec.\ \ref{dspn}, which generates a better depth map $D_r$ via refining the coarse depth map $D_0$.

% \subsection{Loss Function}
During training, we apply $L_2$ loss for the coarse depth output from the prediction network $L_D=\frac{1}{n}\sum_x||D_0(x)-D^*(x)||$.
% \begin{equation}
  % L_D=\frac{1}{n}\sum_x||D_0(x)-D^*(x)||
% \end{equation}
Similarly, we also use $L_2$ loss to supervise the learning of refinement depth $D_r$ and confidence mask $M$. Then the total loss can be written as
\begin{equation}
  L=\lambda L_D + \alpha L_{D_r} + \beta L_M
\end{equation}
where $\lambda$, $\alpha$ and $\beta$ are the weights of the three kind of loss function.

\section{Experiments}
\subsection{Datasets and Experiment setups}
{\bf KITTI Depth completion dataset.}\ The KITTI Depth completion benchmark \cite{uhrig2017sparsity} is a large autonomous driving real-world dataset. It contains 93k depth maps with corresponding raw LiDAR scans and RGB images. 86k of them are separated for training, 7k for validation, and 1k for testing. 

\noindent{\bf Metrics.}\ For evaluation, we adopt the same error metrics in KITTI benchmark, including root mean square error (RMSE), mean absolute error (MAE), Root mean squared error of the inverse depth (iRMSE) and Mean absolute error of the inverse depth (iMAE).

\noindent{\bf Implementation.}\ In our experiments, our model is implemented on PyTorch library and trained for 30 epochs with batch size 16. ADAM optimizer is adopted with the learning rate initialized $1\times 10^{-4}$ and decayed by 0.1 every 10 epochs, and the parameter for weight decay is set to $1\times 10^{-4}$. 
% Specifically, we randomly crop the RGB images and depth maps to $256\times 512$ in training, and the weights $\lambda,\alpha,\beta$ in loss function are set to $1,1,0.1$.

  \begin{figure*}[t]
    \centering
    \includegraphics[width=\textwidth]{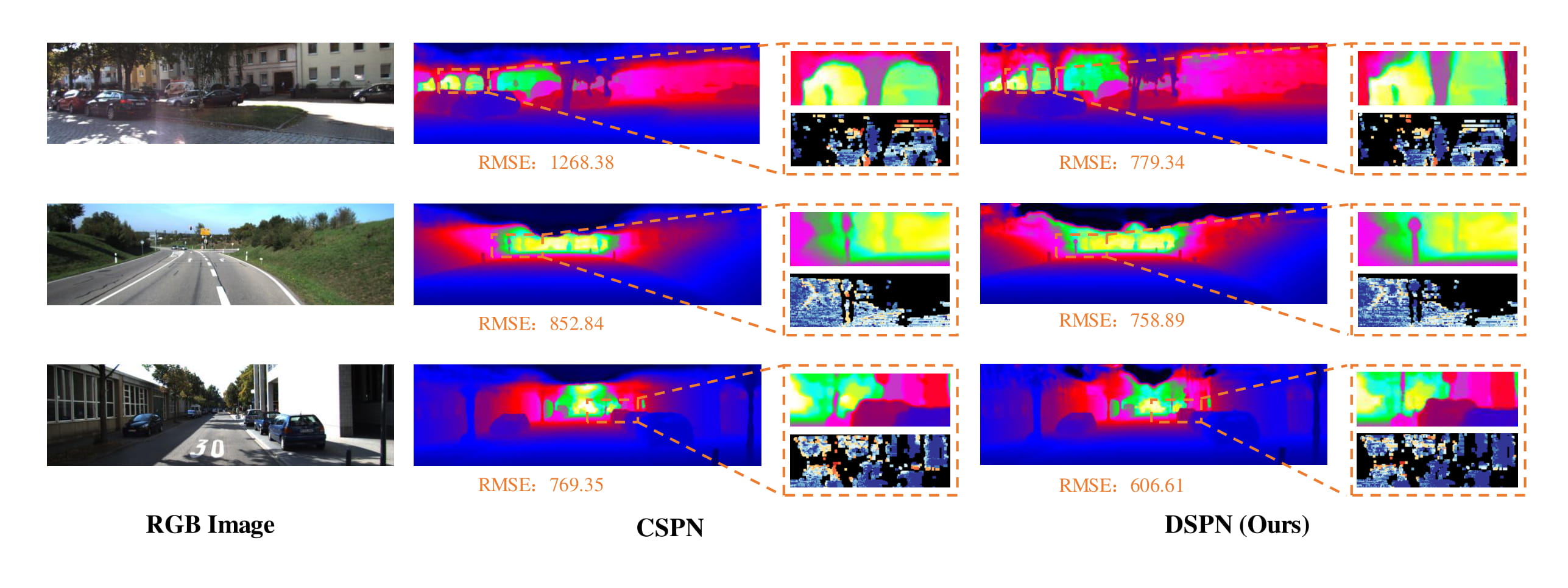}
    \vspace{-0.8cm}
    \caption{Quantitative comparison with CSPN \cite{cheng2018depth} on KITTI test set, where the zoom regions of completion results and error maps show that our method achieves better performance and recover better details.}
    \label{vis}
  \end{figure*}

\subsection{Ablation Study}
To verify the effectiveness of our proposed DSPN, we conduct extensive ablation studies. In detail, we use the prediction network introduced in Sec.\ \ref{net} as baseline in the experiments. Then different variants of refinement network are adopted for comparison, i.e., replacing DSPN with CSPN in our proposed framework and adjusting the parameter settings of CSPN and DSPN. Apart from that, we also investigate the impact of confidence branch. The comparison results are shown in Tab.\ \ref{ablation}, where `Iters' means the numbers of iterations during propagation, and `Size' means the size of receptive field in propagation. Specifically, we can see that (1) the confidence mask can improve the performance by mitigating the measurement errors; (2) As refinement network, our proposed DSPN performances significantly better than CSPN; (3) Compared with CSPN, our proposed DSPN is not sensitive to the number of iterations and the size of receptive field in propagation. These results validate the effectiveness of confidence map and verify that our proposed DSPN only requires far fewer iterative steps and pixels to achieve better performance.

\newcommand{\tabincell}[2]{\begin{tabular}{@{}#1@{}}#2\end{tabular}}
  \begin{table}[t]
    \centering
    \small
    \begin{tabular}{c|cc|cccc}
      \hline
      Method&Iters&Size&RMSE&MAE&iRMSE&iMAE\\
      \hline
      \tabincell{c}{baseline$^*$}&$\backslash$&$\backslash$&825.11&269.44&2.96&1.36 \\
      \hline
      \tabincell{c}{baseline}&$\backslash$&$\backslash$&819.93&257.66&2.91&1.29 \\
      \hline
      \multirow{4}*{\tabincell{c}{baseline \\ + CSPN}}&3&$3\times 3$&816.39&242.61&2.80&1.16 \\
      &6&$3\times3$&814.71&238.42&2.97&1.13 \\
      &12&$3\times3$&813.59&237.00&2.87&1.11 \\
      &12&$5\times5$&810.41&232.15&2.71&1.08 \\
      \hline
      \multirow{3}*{\tabincell{c}{baseline \\ + DSPN}}&3&$3\times 3$&805.90&225.05&2.62&1.05 \\
      &6&$3\times 3$&805.35&223.88&2.56&1.03 \\
      &12&$3\times 3$&804.97&220.32&2.54&1.01 \\
      &12&$5\times 5$&805.23&222.17&2.50&1.01 \\
      \hline
    \end{tabular}
    \caption{The performance comparison of different variants on the validation set of KITTI. baseline denotes the proposed prediction network, $^*$ means without confidence map.}
    \label{ablation}
  \end{table}

\subsection{Comparison with the State-of-the-Arts (SoTA)}
We compare our method with the SoTA methods on the test set of KITTI depth completion benchmark. The comparison results are summarized in Tab.\ \ref{st}, which show that our method achieves SoTA performance on all metrics. Specifically, our method ranks 1st among these mentioned methods according to the RMSE and iMAE metrics. On the other two metrics, PwP\ \cite{xu2019depth} is superior to our method, which may be because it introduces surface normal. Futhermore, we conduct quantitative comparison with the CSPN\ \cite{cheng2018depth} as demonstrated in Fig.\ \ref{vis}. The completion results and error maps in zoom regions show that our method recovers more details and reduce the errors, which verifies the superiority of our proposed method.

\section{Conclusion}
In this work, we propose deformable spatial propagation network (DSPN) to refine depth maps, which can learns offsets and affinity matrices adaptively for better performance in propagation. Compared with previous CSPN, our proposed DSPN can achieve better performance with fewer iterations and pixels in depth completion task. Moreover, we mitigate the effect of noises in LiDAR measurements via introducing a branch to predict the confidence of sparse depth. We further propose a framework to predict dense depth map from a sparse LiDAR depth and a monocular RGB image. Extensive experiments verify the effectiveness of our proposed DSPN and demonstrate that our method achieves the state-of-the-art performance in depth completion task.

\begin{table}[t]
  \centering
  \begin{tabular}{c|cccc}
    \hline
    Method&RMSE&MAE&iRMSE&iMAE\\
    \hline
    DFuseNet\ \cite{shivakumar2019dfusenet}&1206.66&429.93&3.62&1.79\\
    \hline
    CSPN\ \cite{cheng2018depth}&1019.64&279.46&2.93&1.15\\
    \hline
    HMS-Net\ \cite{huang2019hms}&937.48&258.48&2.93&1.14\\
    \hline
    \tabincell{c}{NConv \\ -CNN}\ \cite{eldesokey2019confidence}&829.98&233.26&2.60&1.03\\
    \hline
    \tabincell{c}{Sparse- \\ to-Dense} \ \cite{ma2019self}&814.73&249.95&2.80&1.21\\
    \hline
    PwP\ \cite{xu2019depth}&777.05&{\bf 215.02}&{\bf 2.42}&1.13\\
    \hline
    {\bf DSPN (Ours)}&{\bf 766.74}&220.36&2.47&{\bf 1.03}\\
    \hline
  \end{tabular}
  \caption{Comparison with the state-of-the-art methods on the test set of KITTI depth completion benchmark. The evaluation is done via KITTI testing server.}
  \label{st}
\end{table}

\noindent{\bf Acknowledgment}
\vspace{-2pt}

\noindent This work was partially supported by the National Key Research and Development Program of China (No. 2017YFB13\\02400), the National Natural Science Foundation of China (No. 41571436), and the Hubei Province and Technology Support Program of China (No. 2015BAA027).

\bibliographystyle{IEEEbib}
\bibliography{strings,refs}

\end{document}